\documentclass[letterpaper, 10 pt, conference]{ieeeconf}  

\IEEEoverridecommandlockouts                              

\usepackage{amsmath} 
\usepackage{amssymb}  
\usepackage{amsmath}
\usepackage{algorithm}
\usepackage{algpseudocode}
\usepackage{graphicx}
\usepackage[backref]{hyperref}
\hypersetup{hidelinks}
\usepackage{cuted}
\usepackage{xcolor}
\usepackage{multirow}
\usepackage{tabularx}
\title{\LARGE \bf MagServo: Uncertainty-Resilient Hierarchical \\Magnetic Servoing via Learned Latent Representations

}

\author{
Yuhan Tan$^{1,2,\dagger}$,
Yameng Zhang$^{2,3,\dagger,*}$,
Pei Liu$^{2,4}$,
Yao Zhong$^{2}$,
Zhongliang Jiang$^{2,*}$
\thanks{$^{\dagger}$Yuhan Tan and Yameng Zhang contributed equally to this work.}
\thanks{$^{*}$Corresponding authors: Yameng Zhang and Zhongliang Jiang
(email: yamengzhang@mrc-cuhk.com; zljiang@hku.hk).}
\thanks{$^{1}$Department of Bioengineering, Imperial College London, UK.
$^{2}$Department of Mechanical Engineering, The University of Hong Kong, Hong Kong SAR.
$^{3}$Multi-scale Medical Robotics Center, AIR@InnoHK, Hong Kong SAR.
$^{4}$School of Computation, Information and Technology, Technical University of Munich, Germany.}
}

\begin{document}

\maketitle
\thispagestyle{empty}
\pagestyle{empty}

\begin{abstract}
Magnetic navigation provides contact-free and line-of-sight-independent feedback for robotic systems, yet existing approaches typically rely on explicit pose estimation or direct use of raw magnetic measurements, making accurate control susceptible to modeling errors, measurement noise, and disturbances. This work presents \textbf{MagServo}, a hierarchical learning-based framework for robust 6-DoF magnetic servoing directly using the learned latent magnetic feature. MagServo learns uncertainty-resilient magnetic representations through masked reconstruction and captures state-dependent interaction dynamics between robot motion and latent magnetic transitions without analytical magnetic models or explicit Jacobian supervision. Based on the learned dynamics, a hierarchical controller combines nonlinear model predictive control for coarse approach with local Jacobian inversion for precise fine regulation. Extensive physical experiments demonstrate submillimeter and subdegree accuracy, achieving mean terminal errors of 0.386~mm and 0.479$^{\circ}$ for 6-DoF pose reaching. MagServo further outperforms a localization-based control baseline in complex trajectory tracking and maintains robust performance under unseen magnetic-source configurations without retraining. A supplementary video of the real-robot experiments is available at \url{https://youtu.be/rZt1NUP1Mr0}.

\end{abstract}

\section{Introduction}
\label{sec-1}


\par
Magnetic localization and navigation technologies have attracted increasing attention due to their non-contact, real-time, and line-of-sight-independent nature, enabling reliable operation in visually occluded or confined environments~\cite{wang2026enhancing,xu2022adaptive,li2023closed,huang2026enhancing}. Their radiation-free nature further makes them particularly well suited for medical applications, particularly robot-assisted endoscopic examinations and minimally invasive procedures~\cite{lai2023sim,zhang2023design}. 
In these cases, visual feedback is often compromised due to poor lighting conditions or physical obstructions between the observer and the target, such as a robotic capsule for gastrointestinal examination~\cite{xu2022adaptive,liu2026model}. 

\par
Magnetic navigation methods can be broadly divided into two categories: explicit localization-based navigation and direct raw magnetic data-based servoing. The former requires explicit estimation of the robot end-effector pose from real-time magnetometer measurements using an accurate magnetic model. Navigation is then achieved by minimizing the error between the estimated current pose and the desired pose. To establish the mapping between the magnetic flux density measured by the magnetometers and the robot pose, various magnetic models have been investigated, including the magnetic dipole model~\cite{popek2017first}, analytical model~\cite{zhang2023magnetic}, and complete elliptic integral model~\cite{derby2010cylindrical}. Considering the inherent nonlinearity and modeling uncertainties of this mapping, Son~\emph{et al.} adopted a quadratic measurement model for online calibration~\cite{son2018simultaneous}. To further improve pose estimation accuracy, numerical optimization methods, such as particle swarm optimization~\cite{shi1998modified} and the Levenberg-Marquardt algorithm~\cite{gavin2019levenberg}, are commonly employed to minimize the discrepancy between the measured and model-predicted magnetic fields. While these model-based approaches offer strong physical interpretability, they often require careful system calibration and computationally intensive iterative optimization, limiting their efficiency for real-time applications~\cite{wu2022calibration}.

\begin{figure}
    \centering
    \includegraphics[width=0.95\columnwidth]{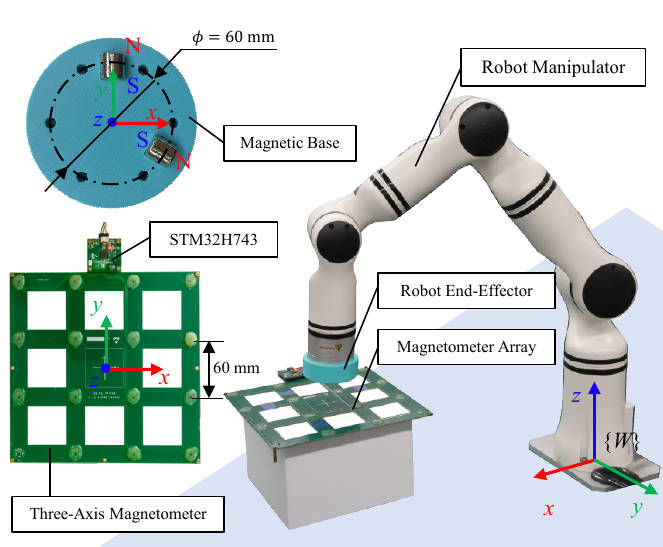}
    \caption{Magnetic servoing system comprising a robot manipulator, a dual-magnet end-effector fixture, and a $4\times4$ magnetometer array.}
    \label{fig:setup}
\end{figure}

\par
Recently, to reduce the complexity associated with parameter identification and iterative optimization while improving localization accuracy, learning-based methods have emerged to directly estimate robot poses from magnetometer measurements. Su~\emph{et al.} proposed AMagPoseNet for 6-DoF pose estimation of an annular magnet, where PoseNet is pretrained using data generated from a mathematical model and CaliNet adapts the learned mapping to the real magnetic field using a small amount of experimental data~\cite{su2023amagposenet}. This model-assisted learning strategy substantially reduces the need for large-scale real-world data collection. Similarly, Zhao~\emph{et al.} employed differential magnetic signals and a precisely calibrated magnetic model to generate large-scale simulated data for training a deep network, enabling accurate and real-time pose estimation~\cite{zhao2025deep}. Yang~\emph{et al.} proposed PRPosNet, which combines CNN-based local feature extraction, coordinate attention, and Transformer-based modeling to capture both local spatial characteristics and long-range dependencies among magnetic sensors~\cite{yang2025prposnet}. Beyond direct pose regression, recent studies have increasingly incorporated physical priors into neural models to improve accuracy and generalizability. Shen~\emph{et al.} proposed a physics-informed residual network with a dual-branch architecture to explicitly correct discrepancies between the ideal magnetic dipole model and real magnetic measurements~\cite{shen2026physics}. To account for measurement noise in practical magnetometers, Xie~\emph{et al.} augmented ideal magnetic data generated from a theoretical model with Gaussian noise and trained PoseNet on the synthetically corrupted measurements, achieving localization errors of approximately 1.54~mm in position and $2.24^{\circ}$ in orientation~\cite{xie2025theoretical}.

\par
Instead of explicitly estimating the robot pose, direct raw magnetic data-based servoing uses magnetometer measurements as feature descriptors and directly actuates the robot end-effector to minimize the corresponding feature errors~\cite{zhang2024model}. This eliminates the need to map magnetic flux density measurements into Cartesian poses. However, direct servoing requires a Jacobian matrix that relates robot motion to variations in magnetometer measurements, which is challenging to derive analytically. To address this issue, Su~\emph{et al.} pioneered an online Jacobian estimation approach, in which the Jacobian matrix is vectorized as a state variable within a state-space formulation and estimated online using a Kalman filter (KF)~\cite{qian2002online}. Building upon this formulation, Tan~\emph{et al.} and Edelmann~\emph{et al.} developed model-free control strategies for magnetically actuated endoscopes~\cite{tan2022model,edelmann2018estimation}. However, the performance of such methods strongly depends on appropriate characterization of the process and measurement noise statistics used by the KF. To relax the reliance on predefined static noise models, adaptive approaches have been introduced to dynamically update the process and measurement noise covariances during online control~\cite{li2017model,yang2022model}. Nevertheless, these approaches primarily estimate the Jacobian from feature variations and robot motion at the current control step, making the estimation susceptible to instantaneous measurement noise and disturbances. To further improve servoing stability, Zhang~\emph{et al.} introduced a Jacobian consistency assumption and incorporated a moving window of historical feature-error and actuation data, which are linearly combined with current observations to achieve more robust Jacobian estimation~\cite{zhang2024model}. However, this approach remains susceptible to environmental disturbances and transient measurement drift induced by sensor noise.

\par
To address these challenges, we introduce \textbf{MagServo}, a hierarchical learning-based framework for robust and accurate magnetic servoing. MagServo learns uncertainty-resilient magnetic representations through masked reconstruction and captures their motion-conditioned dynamics without requiring analytical magnetic models or explicit Jacobian supervision. The representation and dynamics are further jointly refined through multi-step rollout learning, establishing a robust, control-aware latent space for magnetic servoing. Leveraging the learned dynamics, MagServo combines nonlinear model predictive control (MPC) for rapid coarse approach with local Jacobian inversion for precise fine regulation. By directly translating latent magnetic discrepancies into corrective robot motions, MagServo bypasses explicit magnetic-to-pose estimation and enables accurate closed-loop 6-DoF servoing. The contributions of this work are summarized as follows:
\begin{itemize}
    \item We propose a direct magnetic servoing algorithm that uses learned latent magnetic representations as control features, eliminating explicit pose estimation and analytical magnetic modeling while remaining robust to measurement noise and disturbances.
    
    \item We introduce MagServo, which jointly learns latent magnetic representations and state-dependent interaction dynamics and integrates them into a hierarchical coarse-to-fine controller, combining nonlinear MPC with local Jacobian inversion for accurate and reliable magnetic servoing.
\end{itemize}
To evaluate MagServo, we conduct extensive physical experiments on 6-DoF pose reaching, complex trajectory tracking, and unseen magnetic-source configurations to assess its accuracy and generalizability. MagServo achieves submillimeter and subdegree accuracy, with mean terminal errors of 0.386 mm and 0.479$^{\circ}$ for pose reaching. Its effectiveness is further demonstrated by substantially outperforming the localization-based baseline and maintaining robust performance under unseen magnetic-source configurations without retraining.

\begin{figure*}[!tb]
\centering
\includegraphics[width=0.95\linewidth]{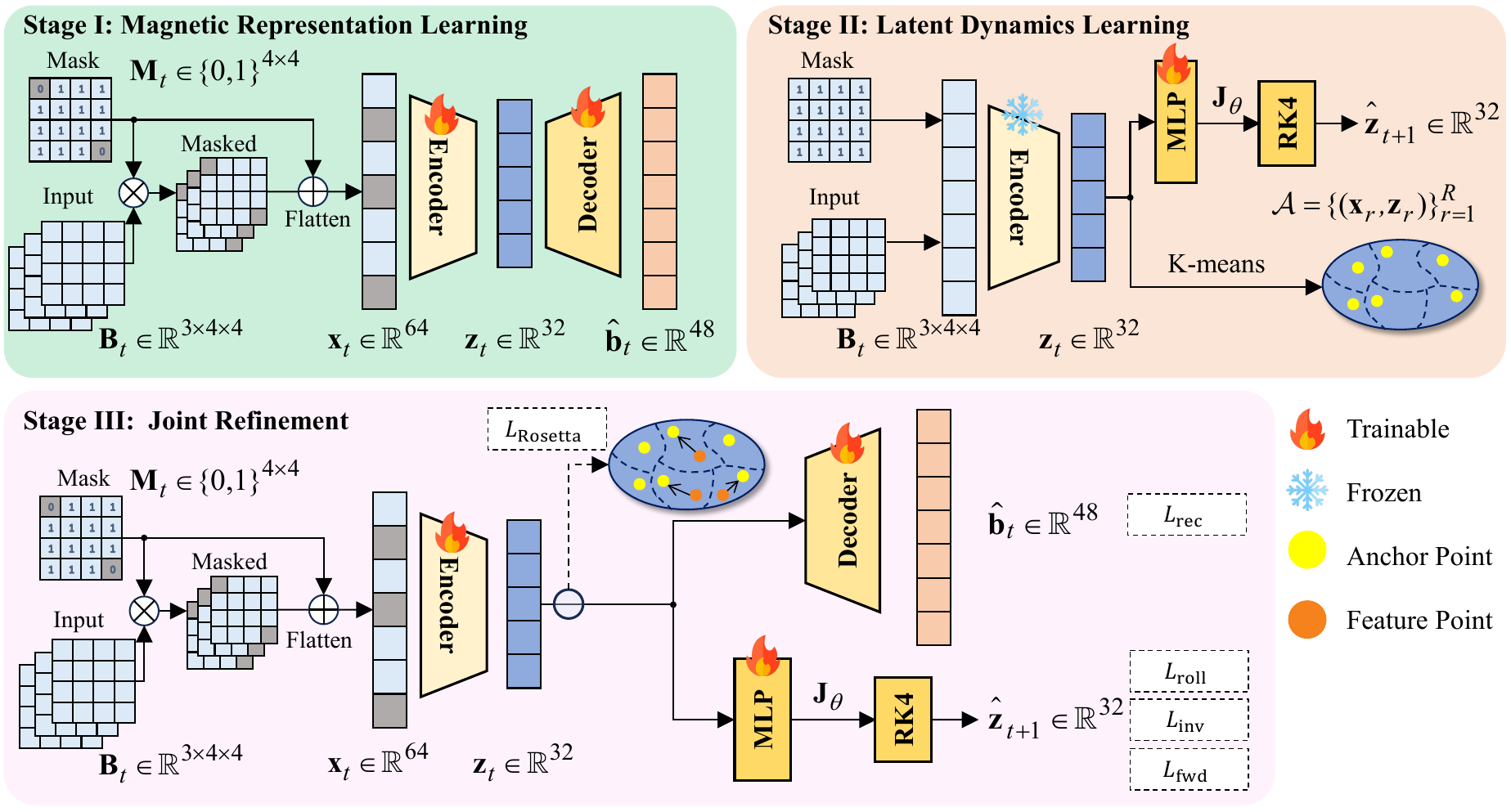}
\caption{Three-stage magnetic representation training. Stage I learns magnetic representations by reconstructing complete observations from masked inputs. Stage II learns latent dynamics with the encoder frozen and selects representative anchors through clustering. Stage III jointly refines the representation and dynamics using reconstruction and prediction losses, with Rosetta anchor regularization to limit latent coordinate drift.}
\label{fig:offline_training}
\end{figure*}

\section{System Overview and Problem Formulation}
\label{subsec:problem}

\subsection{System Configuration}
As shown in Fig.~\ref{fig:setup}, the system couples a robot manipulator carrying two permanent magnets with a stationary $4\times4$ array of three-axis magnetometers. The axially magnetized cylindrical magnets are rigidly attached to the end-effector, with their magnetization axes separated by $120^\circ$ to reduce the rotational ambiguity of a single axisymmetric magnet. As the end-effector translates and rotates relative to the array, the resulting changes in the measured field provide feedback for regulating its motion.

The magnetic measurements are transmitted to a host computer, where the controller computes incremental Cartesian commands for the robot. After each increment is executed, a new observation is acquired to update the next command. This closed-loop interaction enables the robot to approach a desired magnetic observation through successive corrections. The relation between robot motion and magnetic feedback is learned from recorded data, without requiring an analytical magnetic field model.

\subsection{Task Definition}
Our objective is to achieve accurate 6-DoF end-effector positioning and path tracking using magnetic feedback. For single-pose reaching, the target is specified by a magnetic observation $\mathbf{B}^{\star}$ recorded at the desired position and orientation. The task is to reach this configuration from different initial poses. For path tracking, the target is an ordered sequence of magnetic observations $\{\mathbf{B}_k^{\star}\}_{k=1}^{K}$ recorded along a reference path, and the task is to follow the corresponding configurations in sequence. In both tasks, the goal is to minimize physical position and orientation errors while maintaining stable motion within the robot's operating constraints. Accurate reaching and tracking should be retained in the presence of magnetic measurement noise and disturbances. Performance is assessed using pose errors relative to the recorded reference configurations.

\section{Methods}
\label{sec:methods}
The proposed MagServo first trains an autoencoder offline with random masking to learn robust latent representations of magnetic measurements under unknown noise and sensor disturbances. An MLP is then trained to model the latent interaction dynamics, predicting the feature transition from time $t$ to $t+1$ conditioned on the applied robot motion. Finally, the AE and MLP are jointly optimized to improve the consistency between latent representation and interaction dynamics. The overall neural representation learning pipeline is illustrated in Fig.~\ref{fig:offline_training}. In this study, we proposed MagServo using learned latent features directly as control features.


\subsection{Magnetic Neural Representation Learning}
\label{subsec:offline_training}


\subsubsection{Training Data Preparation}
\label{subsec:input_preparation}
The training data comprise robot transitions $(\mathbf{B}_t,\mathbf{u}_t,\mathbf{B}_{t+1})$, where $\mathbf{B}_t\in\mathbb{R}^{3\times4\times4}$ is the magnetic observation. The normalized action is
\begin{equation}
    \mathbf{u}_t=\mathbf{S}_u
    \begin{bmatrix}
        \Delta\mathbf{p}_t^{E}\\
        \Delta\boldsymbol{\phi}_t^{E}
    \end{bmatrix},
    \quad
    \mathbf{S}_u=\operatorname{diag}\!\left(s_p^{-1}\mathbf{I}_3,s_\phi^{-1}\mathbf{I}_3\right),
    \label{eq:executed_action}
\end{equation}
where $\Delta\mathbf{p}_t^{E}$ and $\Delta\boldsymbol{\phi}_t^{E}$ denote the measured translational and rotation-vector increments expressed in the current end-effector frame, respectively, and $s_p,s\phi>0$ are the corresponding normalization scales. Each magnetic observation is then background-corrected, standardized channel-wise, and vectorized as follows:
\begin{equation}
    \mathbf{b}_t
    =
    \operatorname{vec}
    \left(
        \frac{\mathbf{B}_t-\mathbf{B}_e-\boldsymbol{\mu}}
             {\boldsymbol{\sigma}}
    \right)
    \in\mathbb{R}^{48},
    \label{eq:magnetic_preprocessing}
\end{equation}
where $\mathbf{B}_e$ is the recorded background magnetic field, and $\boldsymbol{\mu}$ and $\boldsymbol{\sigma}$ denote the channel-wise mean and standard deviation of the background-corrected training data, respectively, broadcast across all sensors. The same preprocessing procedure is applied to both current and reference magnetic observations.

\subsubsection{Stage 1: Magnetic Representation Learning}

As shown in Fig.~\ref{fig:offline_training}, an AE learns robust magnetic representations by reconstructing complete observations from intact and/or partially masked inputs. A binary mask $\mathbf{M}_t\in{0,1}^{4\times4}$ is applied sensor-wise across all three magnetic field components:
\begin{equation}
    \widetilde{\mathbf{B}}_t=\mathbf{M}_t\odot\mathbf{B}_t,
    \label{eq:magnetic_augmentation}
\end{equation}
where $\odot$ broadcasts the mask across channels. The masked observation undergoes the preprocessing in Eq.~\eqref{eq:magnetic_preprocessing} to obtain $\widetilde{\mathbf{b}}_t\in\mathbb{R}^{48}$. To identify the masked sensor locations, the encoder input includes the vectorized mask:
\begin{equation}
    \mathbf{x}_t=\operatorname{Concat}(\widetilde{\mathbf{b}}_t,\mathbf{m}_t)
    \in\mathbb{R}^{64},
    \quad \mathbf{m}_t=\operatorname{vec}(\mathbf{M}_t).
    \label{eq:augmented_encoder_input}
\end{equation}
The clean inputs use an all-ones mask. The encoder and decoder produce: 
\begin{equation}
    \mathbf{z}_t=E_\phi(\mathbf{x}_t)\in\mathbb{R}^{d_z},
    \qquad
    \widehat{\mathbf{b}}_t=D_\psi(\mathbf{z}_t)\in\mathbb{R}^{48},
    \label{eq:autoencoder}
\end{equation}
where $\mathbf{z}_t$ denotes the learned latent feature used as the state for subsequent dynamics learning, and $d_z$ denotes the latent feature dimension, empirically set to 32 in this study. The reconstruction loss is defined as follows:
\begin{equation}
    \mathcal{L}_{\mathrm{rec}}
    =\operatorname{Huber}(\widehat{\mathbf{b}}_t,\mathbf{b}_t),
    \label{eq:stage1_reconstruction}
\end{equation}

The loss is averaged over the batch and all 48 components of the complete target $\mathbf{b}_t$, including the masked components. This objective preserves information from intact observations while encouraging the recovery of missing measurements, with the Huber loss reducing sensitivity to large residuals.


\subsubsection{Stage 2: Latent Dynamics Learning}
\label{subsec:interaction_model}
In order to learn the local interaction dynamics between robot motion and latent magnetic features, the Stage 1 encoder is frozen and a local interaction matrix is learned from latent transitions $(\mathbf{z}_t,\mathbf{u}_t,\mathbf{z}_{t+1})$. As shown in Fig.~\ref{fig:offline_training}, a residual multilayer perceptron (MLP) maps the latent state to a vector of Jacobian entries, which is then reshaped as: 
\begin{equation}
    \mathbf{J}_\theta(\mathbf{z})
    =\operatorname{reshape}_{d_z\times6}
    \bigl(g_\theta(\mathbf{z})\bigr),
    \label{eq:learned_interaction}
\end{equation}
where $g_\theta:\mathbb{R}^{d_z}\rightarrow\mathbb{R}^{6d_z}$. $\mathbf{J}_\theta(\mathbf{z})$ relates normalized motion to local changes in the latent representation through
\begin{equation}
    \frac{d\mathbf{z}(s)}{ds}
    =\mathbf{J}_\theta(\mathbf{z}(s))\mathbf{u}_t,
    \quad \mathbf{z}(0)=\mathbf{z}_t,
    \label{eq:latent_neural_ode}
\end{equation}
where $s\in[0,1]$ parameterizes the motion increment, with $\mathbf{u}_t$ held constant over the interval. A single fourth-order Runge--Kutta (RK4) integration step then yields the predicted latent state:
\begin{equation}
    \widehat{\mathbf{z}}_{t+1}
    =\Phi_\theta(\mathbf{z}_t,\mathbf{u}_t)
    \approx\mathbf{z}(1).
    \label{eq:latent_rk4}
\end{equation}

RK4 re-evaluates $\mathbf{J}_\theta$ at intermediate latent states~\cite{quan2026particlegs}, thereby accounting for its variation along the motion trajectory. Consequently, the transition function $\Phi_\theta$ depends on both the latent state and action, while the learned interaction matrix $\mathbf{J}_\theta$ remains state-dependent only.
The MLP is trained through the RK4 prediction without ground-truth Jacobian labels. The forward loss is:
\begin{equation}
    \mathcal{L}_{\mathrm{fwd}}
    =\frac{1}{|\mathcal{T}|d_z}
    \sum_{t\in\mathcal{T}}
    \|\widehat{\mathbf{z}}_{t+1}-\mathbf{z}_{t+1}\|_2^2,
    \label{eq:stage2_forward}
\end{equation}
where $\mathcal{T}$ indexes the batch transitions. Inspired by the bidirectional Jacobian-learning objective proposed by Przystupa~\emph{et al.}~\cite{przystupa2021analyzing}, we introduce an inverse-consistency loss to support local motion inversion. Specifically, we penalize the latent prediction residual mapped into action space through a damped pseudoinverse: 
\begin{equation}
    \mathcal{L}_{\mathrm{inv}}
    =\frac{1}{6|\mathcal{T}|}
    \sum_{t\in\mathcal{T}}
    \left\|\mathbf{J}_{\lambda,t}^{\dagger}
    (\Delta\mathbf{z}_t-\mathbf{J}_t\mathbf{u}_t)\right\|_2^2,
    \label{eq:stage2_inverse}
\end{equation}
where $\mathbf{J}_t=\mathbf{J}_\theta(\mathbf{z}_t)$, $\Delta\mathbf{z}_t=\mathbf{z}_{t+1}-\mathbf{z}_t$, and $\mathbf{J}_{\lambda,t}^{\dagger}=(\mathbf{J}_t^\top\mathbf{J}_t+\lambda^2\mathbf{I}_6)^{-1}\mathbf{J}_t^\top$ with $\lambda>0$. Only $\theta$ is optimized using
\begin{equation}
    \mathcal{L}_{\mathrm{dyn}}
    =\mathcal{L}_{\mathrm{fwd}}
    +\beta_{\mathrm{inv}}\mathcal{L}_{\mathrm{inv}},
\end{equation}
where $\beta_{\mathrm{inv}} = 0.01$ weights the inverse-consistency term.

\subsubsection{Stage 3: Joint Optimization}
As shown in Fig.~\ref{fig:offline_training}, Stage 3 unfreezes the encoder and decoder and jointly optimizes them with the dynamics MLP. In addition to the reconstruction and dynamics losses introduced in Stages 1 and 2, we incorporate two additional constraints: a rollout loss to improve multi-step prediction accuracy and a Rosetta anchor loss to prevent latent-space drift during joint optimization.


\textbf{Multi-step prediction:}
For each training window of $H$ actions, the model recursively predicts latent states from $\widehat{\mathbf{z}}_t=E_\phi(\mathbf{x}_t)$:
\begin{equation}
    \widehat{\mathbf{z}}_{t+h+1}
    =\Phi_\theta(\widehat{\mathbf{z}}_{t+h},\mathbf{u}_{t+h}),
    \quad h=0,\ldots,H-1.
    \label{eq:stage3_rollout}
\end{equation}
The rollout loss penalizes deviations from the encoded observations over the full prediction horizon:
\begin{equation}
    \mathcal{L}_{\mathrm{roll}}
    =\frac{1}{|\mathcal{T}|Hd_z}
    \sum_{t\in\mathcal{T}}\sum_{h=1}^{H}
    \left\|\widehat{\mathbf{z}}_{t+h}
    -E_\phi(\mathbf{x}_{t+h})\right\|_2^2,
    \label{eq:stage3_rollout_loss}
\end{equation}
where $\mathcal{T}$ indexes the starting time steps of the windows in the mini-batch, and $|\mathcal{T}|$ is the number of windows. This extends one-step supervision to account for errors accumulated through recursive prediction.

\textbf{Latent coordinate regularization:}
Jointly updating the encoder alters the latent coordinates on which the learned dynamics operate~\cite{anderson2026mlasdi}. To constrain this latent-space drift, we construct a fixed anchor set $\mathcal{A}=\{(\mathbf{x}_r,\mathbf{z}_r\}_{r=1}^{R}$ before joint training to preserve representative input–latent correspondences~\cite{martinez2022reproducible}. Using the validation-selected Stage 2 model, intact training observations are encoded with all-ones masks, and their latent representations are clustered using mini-batch $k$-means. The observation nearest to each cluster centroid is selected as an anchor, where $(\mathbf{x}_r,\mathbf{z}_r)$ denotes the anchor input and its latent representation stored before joint refinement. Both are kept fixed, while the current encoding $E\phi(\mathbf{x}_r)$ evolves during joint optimization.
The anchor loss is
\begin{equation}
    \mathcal{L}_{\mathrm{Rosetta}}
    =\frac{1}{Rd_z}\sum_{r=1}^{R}
    \left\|
    \bigl(E_\phi(\mathbf{x}_r)-\mathbf{z}_r\bigr)
    \oslash\mathbf{s}
    \right\|_2^2,
    \label{eq:rosetta_loss}
\end{equation}
where $\oslash$ denotes element-wise division and $\mathbf{s}\in\mathbb{R}^{d_z}$ contains the per-coordinate standard deviations of the intact training encodings before joint refinement. Its entries are lower-bounded by $10^{-6}$ and held fixed during training. This penalty limits changes in representative latent coordinates while allowing the representation to adapt to dynamics learning. In summary, the final combined optimization objective is: 
\begin{equation}
\begin{aligned}
\mathcal{L}_{\mathrm{joint}}
={}&\lambda_{\mathrm{rec}}\mathcal{L}_{\mathrm{rec}}
+\mathcal{L}_{\mathrm{fwd}}
+\beta_{\mathrm{inv}}\mathcal{L}_{\mathrm{inv}}\\
&+\lambda_{\mathrm{roll}}\mathcal{L}_{\mathrm{roll}}
+\lambda_{\mathrm{Rosetta}}\mathcal{L}_{\mathrm{Rosetta}},
\end{aligned}
\label{eq:stage3_objective}
\end{equation}


\begin{figure}
    \centering
    \includegraphics[width=\linewidth]{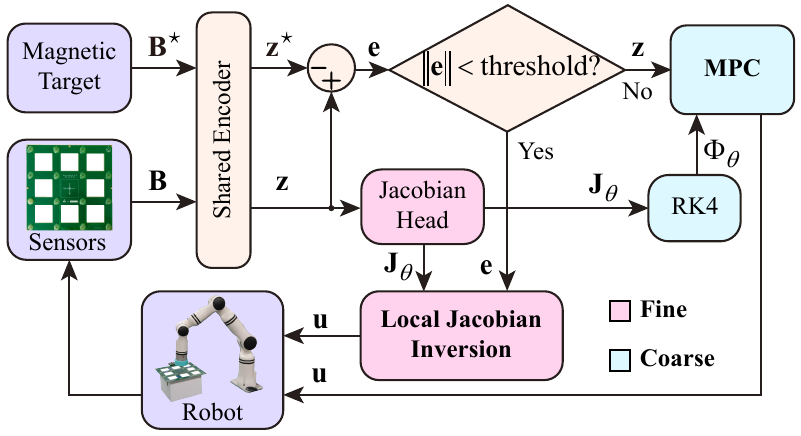}
    \caption{Closed-loop control architecture of MagServo. A shared encoder maps target and measured magnetic fields to latent states. The latent-error norm selects between coarse MPC and local Jacobian inversion. The selected action $\mathbf{u}$ drives the robot, closing the loop through magnetic feedback.}
    \label{fig:control_flow}
\end{figure}

\subsection{Hierarchical Magnetic Servoing}
\label{subsec:online_control}
As shown in Fig.~\ref{fig:control_flow}, the hierarchical controller uses the learned transition model $\Phi_\theta$ for both coarse approach and fine regulation. For large latent tracking errors, model predictive control (MPC) evaluates candidate motion sequences through nonlinear prediction, thereby accounting for variations in magnetic response along the predicted trajectory. Near the reference, small corrective motions permit a local linearization of $\Phi_\theta$ with respect to the action. The fine controller uses this action Jacobian in a constrained damped least-squares problem, avoiding multi-step search while limiting motion in poorly conditioned directions. All the network weights are frozen in this phase.

\subsubsection{Coarse Control via Latent-Space MPC}
At control step $t$, MPC plans $H$ motion increments from the current latent state $\mathbf{z}_t$. Let $k$ index the active reference in the recorded magnetic sequence $\{\mathbf{B}_k^\star\}_{k=1}^{K}$. Its latent encoding $\mathbf{z}_k^\ast$ defines the current tracking error $\mathbf{e}_t=\mathbf{z}_t-\mathbf{z}_k^\ast$. Current and reference observations use the same preprocessing and encoder. For a candidate action sequence $\mathcal{U}_t=\{\mathbf{u}_{t+h}\}_{h=0}^{H-1}$, MPC recursively evaluates the nonlinear transition model:
\begin{equation}
    \widehat{\mathbf{z}}_{t+h+1}
    =\Phi_\theta(\widehat{\mathbf{z}}_{t+h},\mathbf{u}_{t+h}),
    \quad \widehat{\mathbf{z}}_t=\mathbf{z}_t,
    \label{eq:online_nominal_rollout}
\end{equation}
where $h=0,\ldots,H-1$. Each action $\mathbf{u}_{t+h}$ produces a predicted next state with tracking error $\widehat{\mathbf{e}}_{t+h+1}=\widehat{\mathbf{z}}_{t+h+1}-\mathbf{z}_{k+h}^\ast$. The reference sequence starts at the active target and repeats its final entry when needed. MPC selects the action sequence by solving:
\begin{equation}
\begin{aligned}
    \mathcal{U}_t^\star\in{}&
    \underset{\mathcal{U}_t\in\mathcal{F}_t}{\arg\min}\;\Biggl\{
    \sum_{h=1}^{H}\|\widehat{\mathbf{e}}_{t+h}\|_2^2+\lambda_u\sum_{h=0}^{H-1}\|\mathbf{u}_{t+h}\|_2^2\\
    &\quad+\lambda_{\Delta u}\sum_{h=1}^{H-1}
    \|\mathbf{u}_{t+h}-\mathbf{u}_{t+h-1}\|_2^2
    \Biggr\}.
\end{aligned}
\label{eq:coarse_mpc_cost}
\end{equation}
The three terms penalize latent tracking error, action magnitude, and variation between planned actions, respectively, with nonnegative weights $\lambda_u$ and $\lambda_{\Delta u}$. The feasible set $\mathcal{F}_t$ contains action sequences satisfying:
\begin{equation}
\begin{aligned}
    \mathbf{u}_{\min}&\preceq\mathbf{u}_{t+h}\preceq\mathbf{u}_{\max},\\
    \Delta\mathbf{u}_{\min}&\preceq
    \mathbf{u}_{t+h}-\mathbf{u}_{t+h-1}
    \preceq\Delta\mathbf{u}_{\max},
\end{aligned}
\label{eq:mpc_constraints}
\end{equation}
where $\preceq$ denotes component-wise inequality. These bounds restrict each motion increment and its change from the preceding action. At $h=0$, $\mathbf{u}_{t-1}$ denotes the previously requested action and is held fixed during optimization.

The cross-entropy method approximately solves the resulting optimization using predictions from $\Phi_\theta$. Only the first action is executed. After the robot settles, the controller acquires a new magnetic observation and replans, allowing feedback to correct prediction errors at each control step.

\subsubsection{Fine Control via Local Jacobian Inversion}
The fine controller locally linearizes the transition model around zero action at the current latent state. The resulting action Jacobian is given by:
\begin{equation}
    \mathbf{J}_t
    =\left.\frac{\partial\Phi_\theta(\mathbf{z}_t,\mathbf{u})}
    {\partial\mathbf{u}}\right|_{\mathbf{u}=\mathbf{0}}
    =\mathbf{J}_\theta(\mathbf{z}_t).
    \label{eq:fine_action_jacobian}
\end{equation}
The final equality follows from the dynamics in Eq.~\eqref{eq:latent_neural_ode} and the unit-interval RK4 update. At zero action, all intermediate RK4 states remain equal to $\mathbf{z}t$, yielding the stated equivalence. Note that this equivalence holds only at zero action; for a nonzero action, the transition derivative generally differs from $\mathbf{J}_\theta(\mathbf{z}_t)$. Since $\Phi_\theta(\mathbf{z}_t,\mathbf{0})=\mathbf{z}_t$, the predicted tracking error for a small correction is defined as: 
\begin{equation}
    \widehat{\mathbf{e}}_{t+1}
    \approx\mathbf{e}_t+\mathbf{J}_t\mathbf{u},
    \label{eq:fine_linearisation}
\end{equation}
where the active reference is held fixed and $\mathbf{e}_t=\mathbf{z}_t-\mathbf{z}_k^\ast$ denotes the current latent tracking error. Assuming $\mathbf{J}_t$ remains constant within the local update, the corrective action is obtained by solving a constrained damped least-squares problem:
\begin{equation}
\begin{aligned}
\mathbf{u}_t
=\underset{\mathbf{u}_{\min}\preceq\mathbf{u}\preceq\mathbf{u}_{\max}}
{\arg\min}\;\Bigl\{
&\|\mathbf{e}_t+\mathbf{J}_t\mathbf{u}\|_2^2 +\lambda_{\mathrm{damp}}^2\|\mathbf{u}\|_2^2\Bigr\},
\end{aligned}
\label{eq:fine_objective}
\end{equation}
where $\lambda_{\mathrm{damp}}>0$ limits motion along poorly conditioned directions, and the bounds restrict the normalized motion increments.

\subsubsection{Reliability-Aware Mode Switching}
Mode selection combines the latent tracking error with the reliability of the local Jacobian. Define $\epsilon_t=\|\mathbf{e}_t\|_2$, where $\mathbf{e}_t=\mathbf{z}_t-\mathbf{z}_k^\ast$. The controller starts in coarse mode. It enters fine mode when $\epsilon_t<\epsilon_{\mathrm{enter}}$ and returns to coarse mode when $\epsilon_t>\epsilon_{\mathrm{exit}}$. Here, $\epsilon_{\mathrm{enter}}$ and $\epsilon_{\mathrm{exit}}$ are the entry and exit thresholds, chosen such that $0<\epsilon_{\mathrm{enter}}<\epsilon_{\mathrm{exit}}$. Within the interval between them, the current mode is retained. This hysteresis reduces repeated switching caused by fluctuations in magnetic measurements.

A small tracking error alone does not guarantee reliable local inversion. Fine control additionally requires the action Jacobian $\mathbf{J}_t\in\mathbb{R}^{d_z\times6}$, defined in Eq.~\eqref{eq:fine_action_jacobian}, to contain finite entries and satisfy:
\begin{equation}
    \sigma_{\min}(\mathbf{J}_t)\geq\sigma_{\mathrm{floor}},
    \qquad
    \kappa(\mathbf{J}_t)\leq\kappa_{\max},
    \label{eq:jacobian_reliability}
\end{equation}
where $\sigma_{\min}$ and $\sigma_{\max}$ are the smallest and largest singular values, and $\kappa(\mathbf{J}_t)=\sigma_{\max}(\mathbf{J}_t)/\sigma_{\min}(\mathbf{J}_t)$ is the spectral condition number. The positive threshold $\sigma_{\mathrm{floor}}$ excludes rank-deficient or weakly responsive motion directions, while $\kappa_{\max}$ limits ill-conditioning. Failure of either condition, or a nonfinite matrix entry, overrides the error thresholds and selects coarse control. These checks apply both when entering fine mode and while remaining in it.

Following mode selection, each candidate action is validated for motion feasibility and a predicted reduction in the current latent error. If the fine-control candidate is rejected, the controller switches to coarse control using the same observation. Recovery candidates are considered when necessary. If no admissible correction is found, a zero command is issued and the active reference is retained.


\section{Experimental Validation}
\label{sec:experiment}
\par
This section evaluates the proposed method on a physical robotic platform through single-pose reaching, complex trajectory following, and robustness against disturbances introduced by an additional magnetic source. Tracking accuracy is further compared with a magnetic localization-based navigation baseline. Performance is quantitatively assessed in terms of Cartesian pose errors and control efficiency.


\subsection{Experimental Setup}
\label{sec:experiment_setup}
The experimental platform comprises a 7-DoF RealMan 75 manipulator (RealMan Intelligent Technology, China) and a stationary $4\times4$ magnetic sensor array. A 3D-printed end-effector fixture holds two identical N38 NdFeB cylindrical magnets, each $10~\mathrm{mm}$ in diameter and height, with their magnetization axes separated by $120^\circ$ (see Fig.~\ref{fig:setup}). This non-parallel dual-magnet configuration breaks the rotational symmetry of a single axially magnetized cylindrical magnet, thereby reducing ambiguity in 6-DoF pose estimation~\cite{zhang2024model}. The sensor array consists of 16 LIS3MDL tri-axial magnetometers with a spacing of $60~\mathrm{mm}$, a measurement range of $\pm16~\mathrm{Gs}$, and root-mean-square noise levels of $3.2$, $3.2$, and $4.1~\mathrm{mGs}$ along the $x$, $y$, and $z$ axes, respectively. An STM32H743 microcontroller acquires the sensor measurements and transmits them to the host computer via serial communication at up to $20~\mathrm{Hz}$. After preprocessing, the magnetic measurements serve as the feedback signal for tracking throughout all experiments. Robot feedback is used for motion execution and safety monitoring, while the measured TCP poses are used solely to quantify physical position and orientation errors relative to the given reference poses.


\subsection{Performance on 6-DoF Pose Reaching}
\label{sec:single_pose}

\begin{figure}[bt!]
    \centering
    \includegraphics[width=0.5\textwidth]{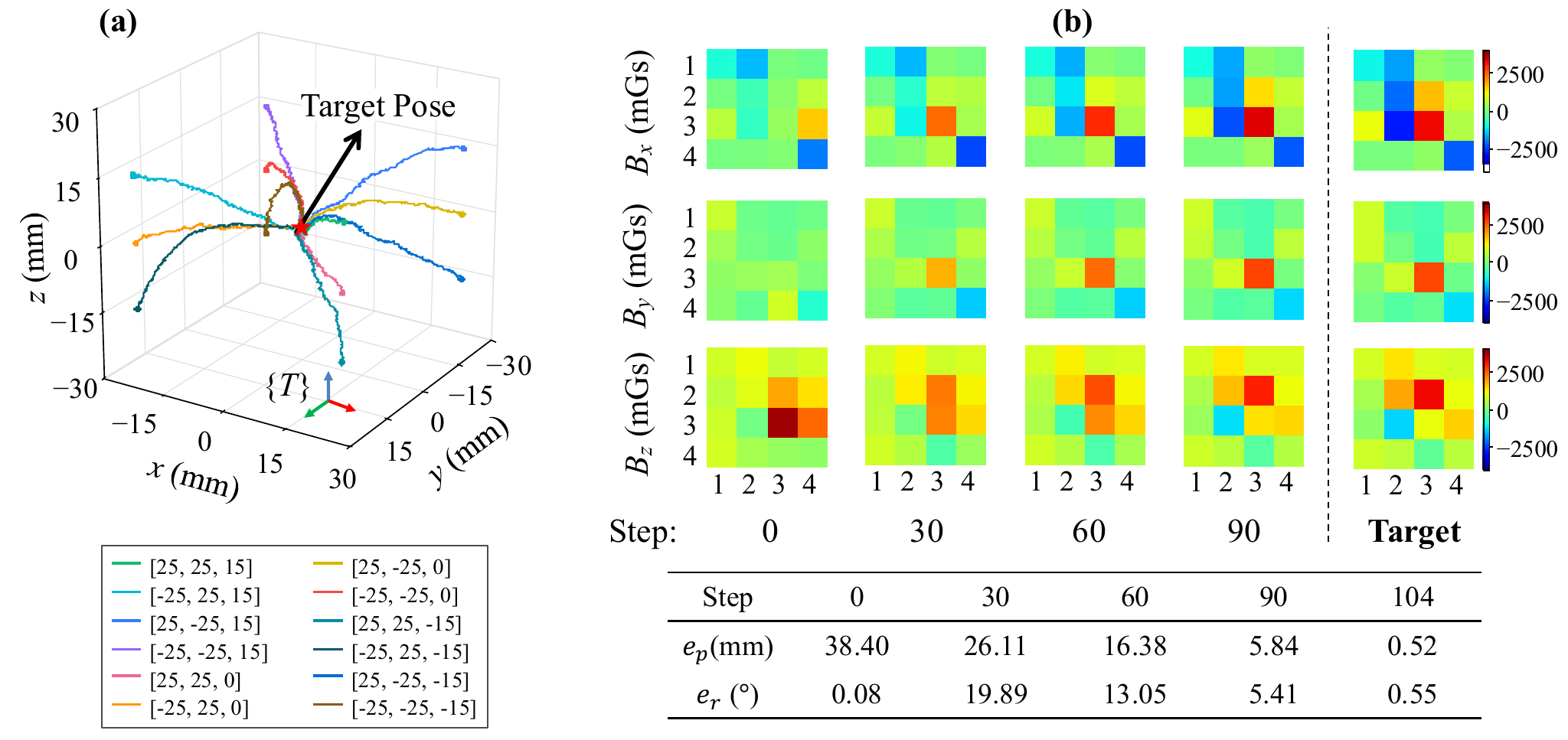}
    \caption{Single-pose reaching results. (a) Servoing trajectories from 12 different initial poses toward the same target pose. (b) Heatmaps of the $4\times4$ magnetic measurements at iterations 0, 30, 60, and 90 compared with the target reference. The table reports the corresponding tracking errors and the final errors at the termination step.    
    }
    \label{fig:single_pose_results}
\end{figure}

To evaluate the performance of the proposed magnetic servoing algorithm, we first conduct a single-pose reaching task. A target pose is arbitrarily selected within the sensing workspace, and the corresponding magnetic measurements are recorded as the reference for servoing. To assess robustness to different initial conditions, 12 trials are performed from different initial poses. These poses combine translational offsets of $\pm25~\mathrm{mm}$ along the $x$ and $y$ axes with offsets of $-15~\mathrm{mm}$, $0$, and $15~\mathrm{mm}$ along the $z$ axis relative to the target pose.

\par
In each trial, the proposed servoing algorithm continuously generates corrective motions toward the target until the predefined latent convergence criterion is satisfied. The control hyperparameters are set to $H=3$, $\epsilon_{\mathrm{enter}}=0.3$, and $\epsilon_{\mathrm{exit}}=0.5$. Fig.~\ref{fig:single_pose_results}(a) visualizes the executed trajectories from different initial poses toward the same target pose, while Fig.~\ref{fig:single_pose_results}(b) shows the magnetic field measurements at selected iterations of a representative trial. As servoing progresses, the measured magnetic field distribution gradually approaches the target distribution, indicating convergence toward the desired pose. As the initial motion is unconstrained, the end-effector may initially deviate from the target; nevertheless, the closed-loop controller progressively corrects the motion and ultimately converges to the target. The final translational and rotational errors converged to $0.52~\mathrm{mm}$ and $0.55^{\circ}$, respectively, after 104 steps.

Across all trials, MagServo successfully reached the target with terminal position and orientation errors below $1~\mathrm{mm}$ and $1^{\circ}$, respectively. The mean terminal position and orientation errors were $0.386 \pm 0.111~\mathrm{mm}$ and $0.479 \pm 0.157^{\circ}$, respectively. These results demonstrate the high accuracy and consistent convergence of MagServo for single-pose reaching from varying initial

\subsection{Performance on Complex Trajectory Following}
\label{trajectory_tracking}

\begin{table*}[t]
\centering
\caption{Tracking accuracy for the selected spiral and raster trajectories.}
\label{tab:tracking_results}
\begingroup
\setlength{\tabcolsep}{4pt}
\renewcommand{\arraystretch}{1.2}
\resizebox{0.85\textwidth}{!}{%
\begin{tabular}{c|ccc|ccc|c}
\hline
Trajectory
& \multicolumn{3}{c|}{$e_p$ (mm)}
& \multicolumn{3}{c|}{$e_r$ ($^\circ$)}
& Actions \\
& $\delta_x$ & $\delta_y$ & $\delta_z$
& $\phi_x$ & $\phi_y$ & $\phi_z$
& \\
\hline
Spiral
& $0.179\pm0.196$ & $0.238\pm0.223$ & $0.055\pm0.048$
& $0.156\pm0.136$ & $0.128\pm0.122$ & $0.084\pm0.069$
& 673 \\
Raster
& $0.255\pm0.311$ & $0.580\pm0.485$ & $0.118\pm0.120$
& $0.673\pm0.440$ & $0.288\pm0.343$ & $0.235\pm0.239$
& 894 \\
\hline
\end{tabular}%
}
\endgroup
\par\vspace{3pt}
\end{table*}

\begin{figure}[tb!]
    \centering
    \includegraphics[width=0.46\textwidth]{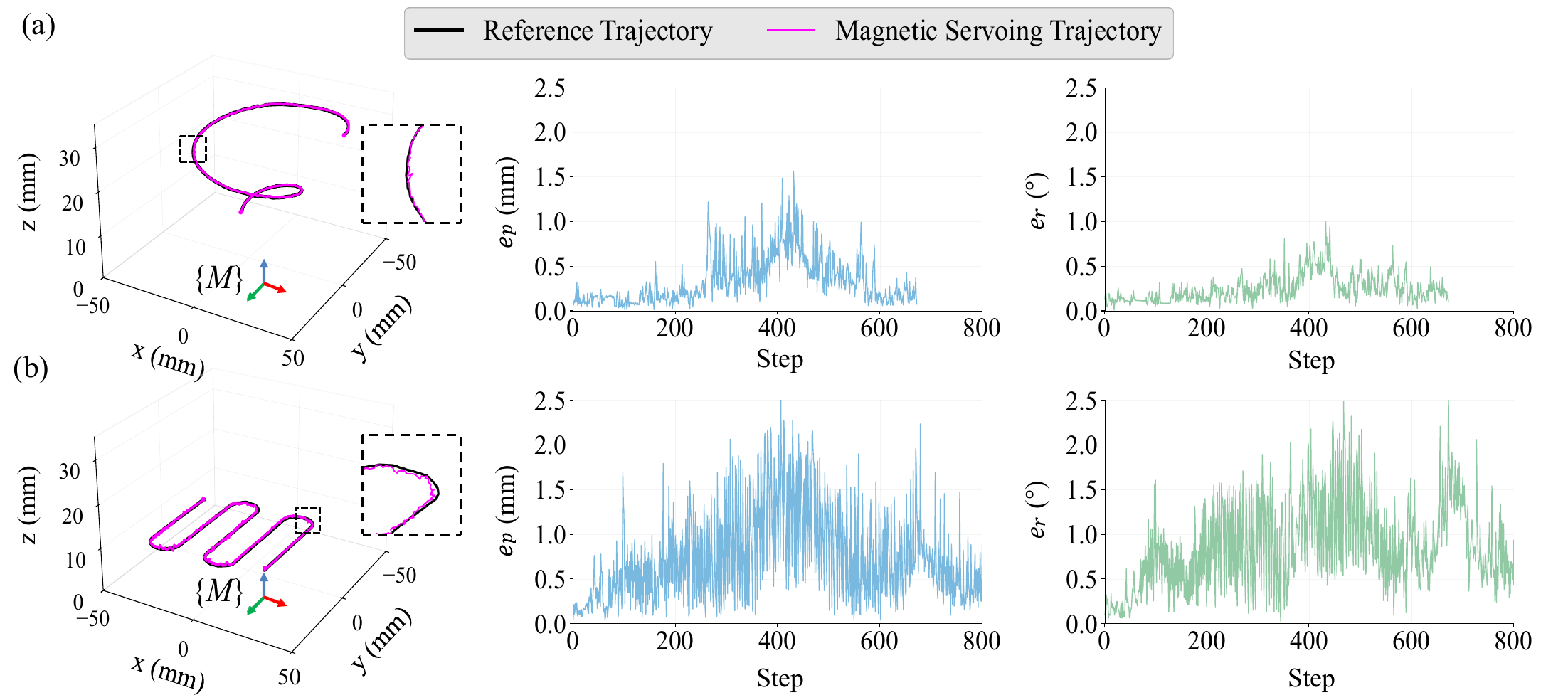}
    \caption{Trajectory-following results. (a) Spiral and (b) raster trajectories. The left column shows the reference and executed servoing trajectories, while the middle and right columns show the corresponding position and rotation errors, respectively. 
    }
    \label{fig:trajectory_tracking}
\end{figure}

\subsubsection{Trajectory Following Results}
Beyond the single-pose reaching task, we further investigate trajectory tracking, which requires continuous correction under changing target configurations and local magnetic responses. To this end, two representative trajectories, spiral and raster, are considered to introduce coupled radial–vertical variations and repeated travel-direction reversals, respectively. The spiral trajectory increases in radius from $10~\mathrm{mm}$ to $45~\mathrm{mm}$ while spanning $24~\mathrm{mm}$ in height. The raster trajectory consists of five parallel straight segments connected by four semicircular U-turns with a radius of $7.5~\mathrm{mm}$, covering a $60\times60~\mathrm{mm}$ region above the sensor array. For each trajectory, a preceding traversal is used to represent the sequence of reference magnetic measurements, which MagServo subsequently tracks through closed-loop servoing.

\par
Experimental results are presented in Fig.~\ref{fig:trajectory_tracking} and Table~\ref{tab:tracking_results}. Across both reference trajectories, MagServo achieved mean absolute errors below $0.6~\mathrm{mm}$ for all translational components and $0.7^{\circ}$ for all rotational components, demonstrating accurate latent-space magnetic servoing across continuously varying reference configurations.

\subsubsection{Comparison Results}
To further demonstrate the effectiveness of the proposed method, we compare MagServo with a localization-based controller that first estimates the end-effector pose from magnetic measurements and then computes corrective motions based on the resulting pose error. Both methods are evaluated on the same robotic platform using the same spiral and raster trajectories. The baseline employs a trained localization network for pose estimation, whereas MagServo directly computes corrective motions from latent magnetic discrepancies using the learned dynamics and local Jacobian.

Both methods successfully completed the reference trajectories. As shown in Table~\ref{tab:baseline_comparison}, MagServo substantially reduced the tracking errors compared with localization-based control. For the spiral trajectory, the position and rotation errors decreased from $1.409~\mathrm{mm}$ to $0.333~\mathrm{mm}$ and from $1.559^{\circ}$ to $0.249^{\circ}$, respectively. Similar improvements were observed for the raster trajectory, with errors reduced from $1.650~\mathrm{mm}$ to $0.712~\mathrm{mm}$ and from $1.538^{\circ}$ to $0.855^{\circ}$. These results demonstrate the advantage of direct latent-space magnetic servoing, which avoids propagating explicit pose-estimation errors into the control loop.

\begin{table}[t]
\centering
\caption{MagServo versus the localization-based baseline.}
\label{tab:baseline_comparison}
\begingroup
\setlength{\tabcolsep}{4pt}
\renewcommand{\arraystretch}{1.2}
\begin{tabular}{cc|cc}
\hline
Trajectory & Method & $e_p$ (mm) & $e_r$ ($^\circ$) \\
\hline
Spiral & Localization & $1.409 \pm 0.673$ & $1.559 \pm 0.627$ \\
       & MagServo & $\mathbf{0.333 \pm 0.267}$ & $\mathbf{0.249 \pm 0.155}$ \\
\hline
Raster & Localization & $1.650 \pm 0.795$ & $1.538 \pm 0.589$ \\
       & MagServo & $\mathbf{0.712 \pm 0.505}$ & $\mathbf{0.855 \pm 0.478}$ \\
\hline
\end{tabular}
\endgroup

\par\vspace{3pt}
\end{table}

\begin{table*}[t]
\centering
\caption{Tracking accuracy with two and three magnetic sources.}
\label{tab:additional_source_tracking}
\begingroup
\setlength{\tabcolsep}{4pt}
\renewcommand{\arraystretch}{1.2}
\resizebox{0.9\textwidth}{!}{%
\begin{tabular}{cc|ccc|ccc|c}
\hline
Trajectory & Magnetic Sources
& \multicolumn{3}{c|}{$e_p$ (mm)}
& \multicolumn{3}{c|}{$e_r$ ($^\circ$)}
& Actions \\
& & $\delta_x$ & $\delta_y$ & $\delta_z$
& $\phi_x$ & $\phi_y$ & $\phi_z$ & \\
\hline
Spiral & Two
& $0.179\pm0.196$ & $0.238\pm0.223$ & $0.055\pm0.048$
& $0.156\pm0.136$ & $0.128\pm0.122$ & $0.084\pm0.069$
& 673 \\
& Three
& $0.304\pm0.245$ & $0.268\pm0.248$ & $0.119\pm0.093$
& $0.253\pm0.176$ & $0.292\pm0.202$ & $0.122\pm0.093$
& 684 \\
\hline
Raster & Two
& $0.255\pm0.311$ & $0.580\pm0.485$ & $0.118\pm0.120$
& $0.673\pm0.440$ & $0.288\pm0.343$ & $0.235\pm0.239$
& 894 \\
& Three
& $0.239\pm0.289$ & $0.491\pm0.489$ & $0.123\pm0.122$
& $0.408\pm0.364$ & $0.226\pm0.285$ & $0.208\pm0.223$
& 824 \\
\hline
\end{tabular}%
}
\endgroup

\par\vspace{3pt}
\end{table*}



\subsection{Performance on Unseen Magnetic-Source Configurations}
\label{sec:additional_mag}

Magnetic servoing should remain robust to magnetic-source configurations unseen during training, as magnetic field characteristics may vary across locations, over time, or due to environmental disturbances. To enhance this robustness, sensor readings are randomly masked during AE training in this study, encouraging the network to reconstruct incomplete magnetic observations and learn latent representations resilient to such variations. To evaluate this capability, an additional magnet, identical to the original two, was mounted on the robot's sixth joint, altering both the magnetic field distribution and its response to robot motion. The same spiral and raster trajectories were then tested using newly acquired reference sequences, while the learned AE model remained unchanged.

As shown in Table~\ref{tab:additional_source_tracking}, MagServo successfully completed both trajectories under the unseen three-source configuration. For the spiral, the additional source moderately increased the tracking errors, while requiring a comparable number of actions (684 vs. 673). Interestingly, the raster trajectory exhibited lower errors in most components and required fewer actions (824 vs. 894), indicating that the additional source does not necessarily degrade servoing performance. Overall, all mean absolute errors remained below $0.50~\mathrm{mm}$ and $0.41^\circ$ in the three-source setting. These results demonstrate that MagServo can accommodate substantial changes in magnetic-source configuration without retraining, while maintaining submillimeter and subdegree navigation accuracy.

\section{Conclusion}
\label{sec-4}
In this work, we presented \textbf{MagServo}, a hierarchical learning-based framework for robust and accurate 6-DoF magnetic servoing. Rather than relying on explicit pose estimation or analytical magnetic models, MagServo directly uses uncertainty-resilient latent magnetic representations as control features. A three-stage learning strategy progressively learns robust magnetic representations and state-dependent interaction dynamics and jointly refines them for closed-loop control. Leveraging the learned dynamics, hierarchical control combines nonlinear MPC for coarse approach with local Jacobian inversion for precise fine regulation. Extensive physical experiments demonstrated submillimeter and subdegree accuracy, with mean terminal errors of $0.386~\mathrm{mm}$ and $0.479^{\circ}$ for 6-DoF pose reaching. MagServo further outperformed localization-based control in complex trajectory tracking and maintained robust performance under unseen magnetic-source configurations without retraining. These results demonstrate the effectiveness of learned latent magnetic representations for accurate, robust, and generalizable magnetic servoing.






\bibliographystyle{IEEEtran}
\bibliography{IEEE}

\end{document}